\documentclass[11pt]{article}
\usepackage[final]{acl}
\usepackage{times}
\usepackage{latexsym}
\usepackage[T1]{fontenc}
\usepackage[utf8]{inputenc}
\usepackage{microtype}
\usepackage{inconsolata}
\usepackage{booktabs}
\usepackage{amsmath}
\usepackage{graphicx}

\DeclareUnicodeCharacter{0169}{\~{u}} 
\DeclareUnicodeCharacter{0129}{\~{\i}} 
\DeclareUnicodeCharacter{1EBD}{\~{e}} 
\DeclareUnicodeCharacter{1EF9}{\~{y}} 

\title{The Constitution Speaks Nheengatu: An Open MT System and Reproducible Corpus Pipeline for the Amazonian Língua Geral}

\author{
  \textbf{Fabrício Carraro\textsuperscript{1}} \quad
  \textbf{Rodolfo Joel Zevallos\textsuperscript{2}} \quad
  \textbf{Rodrigo Gonçalves de Souza\textsuperscript{3}} \\
  \textbf{Caio Henrique Faustino da Silva\textsuperscript{4}} \quad
  \textbf{John E. Ortega\textsuperscript{5}} \\[2pt]
  \normalfont\normalsize
  \textsuperscript{1}Universidad Internacional Menéndez Pelayo \quad
  \textsuperscript{2}Universitat Pompeu Fabra \\
  \normalfont\normalsize
  \textsuperscript{3}Tribunal Regional Federal da 1\textsuperscript{a} Região \quad
  \textsuperscript{4}PUC-Rio \quad
  \textsuperscript{5}Northeastern University
}

\begin{document}
\maketitle

\begin{abstract}
Nheengatu (yrl), the Amazonian Língua Geral, has 6--10 thousand speakers
and is co-official in São Gabriel da Cachoeira (Brazil), yet remains
poorly served by evaluated Portuguese--Nheengatu MT resources. We present the largest parallel corpus assembled for the language (14.5k Portuguese--Nheengatu
pairs), anchored by the 2023 official translation of the Brazilian Federal
Constitution; the first MT evaluation set built from transcribed
spontaneous speech; and, to our knowledge, the first open-weight MT
models dedicated to bidirectional Portuguese--Nheengatu translation.

Three findings organize the paper. First, translation into Nheengatu on
held-out constitutional articles plateaus on development data regardless
of training length or model size, and much of the remaining error sits
in conventional legal terms that the written sources do not attest
consistently; a terminology success rate over a frozen lexicon shows that
glossary mechanisms at training time and lexical constraints at decoding
each raise term realization, the latter at a cost in repetitive output
that chrF++ barely registers. Second, a corpus expansion that adds
transcribed conversation raises speech chrF++ by 24--27 points while
leaving written domains unchanged, a register gap that written-only
benchmarks cannot see. Third, a larger NLLB model helps translation into Portuguese but not into Nheengatu on legal or speech text. 
For Portuguese-to-Nheengatu translation, our final systems reach 31.7--32.8 chrF++ on article-disjoint constitutional text, 64.1--64.5 on the near-domain set (62.1 after removing items with target-side content overlap), and 45.2--47.5 on speech.
\end{abstract}

\section{Introduction}
Machine translation nominally covers hundreds of languages, and the most
recent effort claims more than sixteen hundred \cite{omt2026}, which is under a
quarter of the world's seven thousand \cite{joshi-etal-2020-state}. However, appearing on a list is not the same as being served. The training signal is typically a few thousand sentences of scripture, the evaluation set is a benchmark the language does not appear in, and speakers are rarely asked whether the output is worth reading.

Nheengatu (yrl) is a sharp case of that gap, and an unusual one. Its
ancestor is Tupinambá, the coastal language Europeans met in the
sixteenth century. Jesuit missions carried it up the Amazon as the
language of catechism, where it nativised into the \emph{Língua Geral
Amazônica (Amazonian General Language)} and became the dominant lingua franca of a highly multilingual basin, spoken across ethnic lines by Indigenous and non-Indigenous people alike \cite{cruz2011nheengatu}. It was, for roughly two centuries, the language in which the Amazon conducted its business.

\begin{figure}[t]\centering
\includegraphics[width=\columnwidth]{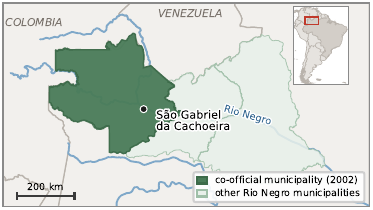}
\caption{The Upper Rio Negro. Dark: São Gabriel da Cachoeira, where
Nheengatu has been co-official since 2002. Boundaries: IBGE; base map:
Natural Earth.}
\label{fig:mapa}
\end{figure}

The Directorate of 1757 then imposed Portuguese and banned the general
language, and nineteenth-century upheavals (the Cabanagem revolt, the
rubber boom, the Paraguayan War) displaced or killed much of the
speaking population. Nheengatu survived chiefly in the Upper Rio Negro
(Figure~\ref{fig:mapa}), with an estimated 6--10 thousand speakers today
\cite{cruz2011nheengatu,quinto2025nheengatu}. Since 2002 it has been
co-official in São Gabriel da Cachoeira, a mostly Indigenous
municipality of 52{,}000 where UFAM trains schoolteachers in the
language.

Nheengatu is a Tupi-Guarani language \cite{cruz2011nheengatu}. It is typologically SVO, agglutinating, postpositional,
with person and number marked on the verb by prefixes and plural by the
particle \emph{ita}, which is usually
written bound in our sources (\emph{mira-itá} `people')
\cite{cruz2011nheengatu,de-alencar-2024-universal}. The morphology is concatenative and moderate rather than polysynthetic.

The yrl side of our corpus holds 14{,}888 distinct forms against 12{,}393 in the matched Portuguese, with 58\% occurring exactly once, so
word-level $n$-grams rarely recur, which is why we report chrF++ and
spBLEU \cite{goyal-etal-2022-flores} rather than word-level BLEU.

Written Nheengatu has no single settled standard: at least four
orthographic traditions circulate in our sources (the academic norm
of Navarro and Avila that the UD treebank follows, the missionary
orthography of the New Testament, the Rio Negro school norm in which the
Constitution was translated, and a convention devised for a smartphone
localisation project) and \citet{dangelis2023orthographies} finds
none better founded than the others. Our transcribed speech adds a
fifth, marking diacritics differently from all of them. A Nheengatu
Language Academy was created in 2021; this paper treats no current
tradition or institution as uniquely authoritative.

Ignoring diacritics, 13.9\% of yrl word types in our training data have two or more realisations, and those types cover 70.3\% of running text; the same measurement on the matched Portuguese side returns 1.7\% and 31.8\%. That the variation is orthographic and not lexical is visible inside a single document: the
Constitution alone writes \emph{istadu-ita} 63 times and \emph{istadu-itá}
23, \emph{deputadu-ita} 10 against \emph{deputadu-itá} 30. Corpus-wide, 1{,}458 sentences write the plural particle bound, 298 write it free, and 48 do both. A system must emit some orthography, and whichever it emits will be unfamiliar to part of its readership; which norm to produce is therefore a community decision, and our contribution is to build the mechanism that can serve whichever
decision is taken.

That co-officiality has no technological counterpart: as of August 2026
the language is absent from NLLB-200, FLORES+, MADLAD, Google Translate
and the 1{,}600-language release of \citet{omt2026}, and to our
knowledge it has no public MT benchmark. The multilingual open-weight
release OmniTranslate lists yrl among its languages but reports no
yrl-specific evaluation \cite{omnitranslate2026}. The prior published
Portuguese--Nheengatu MT results we discuss come from closed industrial
research \cite{goncalves-etal-2025-exploring}.

Two circumstances make this the moment to build one. In 2023 Brazil's
Federal Supreme Court and National Council of Justice published the
Federal Constitution in Nheengatu, translated over two years by fifteen Indigenous translators under editorial review. It is a contemporary, long, non-religious text of a kind that projects at this resource level almost never have. And enough other
material exists (dictionaries, a treebank, school books, transcribed
conversation) that a usable corpus can be assembled if provenance is
tracked source by source.

\paragraph{Contributions.} \textbf{(1) Corpus} (\S\ref{sec:corpus}):
14.5k pt--yrl aligned pairs with source provenance and orthographic metadata, twice the closed corpus of
\citet{goncalves-etal-2025-exploring}, together with article-disjoint legal, pair-deduplicated near-domain, and speech evaluation sets with speakers and sessions held out from training. \textbf{(2) Models} (\S\ref{sec:finaltest}): to our knowledge, the first open-weight MT models dedicated to bidirectional Portuguese--Nheengatu translation. \textbf{(3) Terminology,
diagnosed and treated} (\S\ref{sec:terminology}): a terminology success rate (TSR: the share of glossary terms present in the source that the output realizes), scored over a lexicon frozen before the terminology experiments, identifies unstable conventional terminology as an important contributor to the legal plateau; annotation and token-pair augmentation improve term realization; constrained decoding raises surface-form realization at a 3.2--4.8$\times$ rise in degeneration (repetitive, runaway output), a cost chrF++ barely registers. \textbf{(4) Register vs.\ model size} (\S\ref{sec:register}, \S\ref{sec:scale}): the corpus expansion strongly improves speech while leaving written-domain scores nearly unchanged; within the tested NLLB family, scaling improves yrl$\to$pt and near-domain pt$\to$yrl but not article-disjoint legal or speech pt$\to$yrl in our runs.

\section{Related Work}
\textbf{MT for yrl (closed).} IBM Research Brasil fine-tuned NLLB-600M on
$\sim$6.7k manually cleaned pairs, reporting 69.1/66.9
(pt$\rightarrow$yrl / yrl$\rightarrow$pt) on a 233-segment test of short
dictionary sentences \cite{goncalves-etal-2025-exploring}. Those figures
are \emph{sentence-level} chrF, with large reported standard deviations of $\pm$33.8 and $\pm$34.6; they are not corpus-level chrF++, and they are computed on a private test set; we therefore treat them as context, not as a baseline we can measure a gap against (\S\ref{sec:finaltest}). Their corpus and models are
unpublished, and the same group's human evaluation of those
translators found automatic metrics and speaker-judged usefulness to
diverge substantially \cite{pinhanez-etal-2024-human}, one reason we
treat our own chrF++ gains as necessary but insufficient evidence.

\textbf{Low-resource and Indigenous MT.} No shared task has ever
included Nheengatu: neither WMT
\cite{kocmi-etal-2024-findings,kocmi-etal-2025-findings} nor any of the
five AmericasNLP editions covers it, and it is absent from the 202
languages of NLLB-200, whose only Tupian language is Guarani.
Because no benchmark contains this language, there is no system we can
rank against, and we make no ranking claim anywhere in this paper; what
counts as a meaningful difference in any case shifts with direction,
domain and system proximity \cite{kocmi-etal-2024-thresholds}.

\textbf{Choice of base model.} Five years of AmericasNLP findings
converge on one recipe: fine-tuned NLLB \cite{nllb2022,nllb2024nature}
remains the strongest system for translation \emph{into} Indigenous
languages
\cite{mager-etal-2021-findings,ebrahimi-etal-2023-findings,ebrahimi-etal-2024-findings,de-gibert-etal-2025-findings,stap-araabi-2023-chatgpt},
to the point where the 2025 edition reports a fine-tuned NLLB baseline
that a single team beat, for a single language, by 0.82 chrF++
\cite{de-gibert-etal-2025-findings}.
Grammar-book prompting works mainly for the reverse direction and for languages with richer descriptive material
\cite{tanzer2024mtob,aycock2025grammarbook,pei-etal-2025-understanding,zhang-etal-2024-teaching,zhang-etal-2024-hire},
and post-correction pipelines inherit that asymmetry
\cite{hus-etal-2025-machine}; in-context and modular adaptation of large
models assume retrieval pools or per-language data far beyond what yrl
has \cite{moslem-etal-2023-adaptive,xu2025xalma}. Our own LLM
post-editing experiment is a documented null (\S\ref{sec:decoding}), and
our same-corpus architecture control is mixed once checkpoint selection is equalized (\S\ref{sec:arch}).

\section{The Corpus}
\label{sec:corpus}

\begin{table}[t]\centering\small
\setlength{\tabcolsep}{3pt}
\resizebox{\columnwidth}{!}{%
\begin{tabular}{lrrrl}
\toprule
source & pairs & yrl w.\ & w/s & register \\
\midrule
Constitution 2023 & 2{,}031 & 42{,}262 & 20.8 & legal \\
Avila 2021 (dict.) & 2{,}745 & 24{,}437 & 8.9 & dictionary ex. \\
UD treebank & 2{,}484 & 19{,}270 & 7.8 & mixed \\
LEETRA (4 books) & 526 & 8{,}284 & 15.7 & school \\
Melgueiro 2022 & 1{,}870 & 8{,}128 & 4.3 & dictionary ex. \\
Gurgel (compil.) & 1{,}562 & 7{,}538 & 4.8 & teaching \\
Refubium (speech) & 1{,}156 & 4{,}474 & 3.9 & conversation \\
Tycho Brahe & 441 & 4{,}269 & 9.7 & historical \\
Navarro (course) & 187 & 966 & 5.2 & teaching \\
Catechism 1944 & 23 & 276 & 12.0 & religious \\
\midrule
\textbf{pt--yrl training} & \textbf{13{,}025} & \textbf{119{,}904} & 9.2 & \\
\midrule
\multicolumn{5}{l}{\emph{held out evaluation sets}} \\
legal dev / test & 233 / 218 & 4{,}421 / 3{,}475 & 17.5 & legal article-disjoint \\
near-dom.\ dev / test & 372 / 371 & 2{,}834 / 2{,}701 & 7.4 & dictionary \\
speech dev / test & 116 / 115 & 487 / 537 & 4.4 & conversation \\
\bottomrule
\end{tabular}}
\caption{Corpus composition, ordered by yrl words. ``yrl w.'' is running
words on the Nheengatu side; ``w/s'' is the mean number of words per aligned segment; speech units are pause-delimited turns rather than sentences. Religious text is \textbf{0.2\%} of the corpus (23 pairs), where for most
of the world's languages scripture is the only large digital resource;
our anchor is instead a contemporary legal text translated under
editorial review. }
\label{tab:corpus}
\end{table}

\subsection{Constitution pt--yrl (anchor)}
In 2023 the Brazilian Supreme Court and the National Justice Council
published the Federal Constitution in Nheengatu
\cite{cnj2023nheengatu}, translated over two years by a team of Indigenous translators of the Upper Rio Negro and the Middle Tapajós. It is
to our knowledge the largest modern text in the language, and the only one that is
simultaneously long, contemporary, and translated by speakers under
editorial review. It is also awkward as a corpus: a PDF of a printed
book with no sentence alignment, facing an official Portuguese edition
that has its own amendment history.

We align the two by \emph{document-structure identity} rather than by
length or similarity. Both editions carry the same legal skeleton
(article, paragraph, \emph{inciso} and \emph{alínea}: the numbered item and lettered sub-item of Brazilian legal drafting), so we parse each side with a state machine that validates identifier sequences as it walks the document
(article 5 must be followed by article 6; item XII by item XIII),
and pair units whose full identifier path matches. Structure-based
alignment is what makes the result auditable: every pair carries the
legal address that produced it, so any error is locatable rather than
statistical. Alignment yields 2{,}496 pairs. The final filtered corpus retains
2{,}482: 2{,}031 for training, 233 for development and 218 for testing
(Table~\ref{tab:corpus}).

We verify the Portuguese side against a third source, the Planalto reference edition (Appendix~\ref{sec:repro}). The exercise also surfaced errata in the printed Nheengatu book:
mislabeled items, cascade renumberings, and an entire
constitutional-amendment block absent from the translation. We detected
these precisely because the structural addresses disagreed with the
text. Renumberings we repaired by content; six pairs with genuine
content errors we excluded from the evaluation splits instead of silently
correcting them. Development and test splits hold out \emph{whole
articles} (19 and 19), not sentences, so that no article contributes to
both training and evaluation, with overlap verified to be zero on both
sides.

\subsection{The aggregate corpus and its versions}
Every pair in Table~\ref{tab:corpus} carries its provenance and the
orthographic tradition it was written in: a pair's spelling is only interpretable if one knows which
of the competing norms produced it, and any later normalization decision
must be reversible. Three corpus versions are named in what follows.
v1 is the aggregate as first assembled. v2 applies an audited cleaning
pass removing dictionary apparatus, ellipsis markers and typographic
ALL-CAPS; it changes 170 evaluation pairs, so numbers from the v1
series are never point-compared with later ones. v3 adds 3{,}408 audited
training pairs: (a) \textbf{transcribed spontaneous speech}; (b) \textbf{dictionary example sentences}
(audited against page sources: zero cross-pairings); (c) \textbf{school books} from two
previously uncovered dialect areas, where length-based alignment
\emph{failed} under free translation (one book places translations
\emph{before} the
source stories) and was replaced by semantic realignment with literal
anti-hallucination validation (99.6\% pass). Speech evaluation holds out
an entire session group: 116/115 content-deduplicated, pause-delimited
turns from its two speakers across four elicitation task types, disjoint
from the nine speakers in training; the speech test set is single-use.
The near-domain sets are exact-pair-deduplicated only (not source- or
document-disjoint) and the Limitations quantify what that leaves.
Audits of 200 speech pairs and 240 dictionary-example pairs approve
91.5\% and 86.7\%, respectively. The residual issue classes account for
1.0\% and 6.7\% of the audited samples. The sampling procedure was not
preserved, so these figures do not establish defect rates for the
unaudited remainders. These are not one minus the approval rate, because the two
non-ok tails have different fates; Appendix~\ref{sec:repro} gives the
categories and the derivation. Neither residual is mispairing, which is
tolerable in training data and would not be in evaluation data.

Training additionally includes 925 FR--yrl pairs (a Nheengatu translation
of \emph{Le Petit Prince}, 12{,}467 yrl words): the Nheengatu side is
human, the source side French. They are present in every run reported here. A Portuguese view of the same targets, translated
from the French by a frontier LLM (Claude Fable 5), is a separate
experiment, reported and rejected in \S\ref{sec:terminology}.

\section{Experimental Setup}
\label{sec:setup}
We fine-tune NLLB-200-distilled at 600M and 1.3B
\cite{nllb2022,nllb2024nature}, using the 600M model for replicated configuration comparisons and the 1.3B model for single-seed scale comparisons. Nheengatu is not explicitly supported by NLLB-200 and has no dedicated language token, so we add \texttt{yrl\_Latn} and initialize it from the Guarani \texttt{grn\_Latn} embedding. Training is bidirectional, so a single model serves both directions.

\paragraph{Training glossary.} Two mechanisms are trained jointly. Inline
annotation follows target-lemma annotation
\cite{bergmanis-pinnis-2021-facilitating}: a source term is followed by
its yrl equivalent in a delimited slot (\texttt{term | yrl |}), applied
to 30\% of matched term occurrences in training and, at \emph{hinted}
inference, to every source-side glossary match (the reference is never
consulted), capped at two matches per segment. Token pairs follow the
lexical-augmentation recipe \cite{jones2023bilexrx}: the glossary's
1{,}027 surface forms injected as standalone training pairs. Both draw
on the same audited glossary of 904 entries, mined by Dice association
and adjudicated against source evidence. The glossary is built only from
training pairs and external lexical resources; evaluation references are
never used to construct or adjudicate it.

\paragraph{Scoring lexicon and Terminology Success Rate.} TSR follows the terminology success
rate of the WMT terminology task \cite{semenov-etal-2023-findings}: the
proportion of scoring-lexicon term occurrences in a source sentence whose yrl equivalent (or a listed synonym) appears in the output. The scoring lexicon contains the training glossary. We freeze scoring lexicon v1 \emph{before} the terminology experiments by taking the union of the audited training glossary and the base bilingual lexicon. Every TSR value in this paper uses the same frozen scoring lexicon. A hit counts exact, lemma-table, accent-insensitive and fuzzy matches alike, the last at 0.85 and only for forms of five characters or more. Under this protocol the legal dev set carries 198 term occurrences and the legal test set 232. TSR remains conservative:
\S\ref{sec:analysis} shows by per-case verification that it
\emph{understates} our mechanisms, because inflected, respelled and
fused realizations still count as misses.

\paragraph{Metrics.} We report chrF++ as the primary metric and do not
use a learned metric as primary evidence, because COMET, CometKiwi and
MetricX have not been validated for yrl. We attach no usability label to
any absolute chrF++ value.

\paragraph{Decision discipline.} Configuration decisions use full-dev evaluation and require three seeds; the noise ruler on the legal dev set
is $\pm0.35$ chrF++. We adopted the three-seed rule after two single-seed results failed to replicate, and we report both. Test sets are pre-specified and decoded once per model; \S\ref{sec:finaltest} states which are pristine and which are reused holdouts. A 20k-step 600M run costs $\approx$1.3 GPU-hours, which is what makes three seeds per configuration affordable. Every configuration decision here is
reproducible on a single GPU, and only the 1.3B confirmations need a full node. All $\pm$ values are sample standard deviations over training seeds.

\section{Results}
\subsection{Phase 1: configuration and scale effects}
\label{sec:phase1}
\begin{table}[t]\centering\small
\begin{tabular}{lcccc}
\toprule
 & \multicolumn{2}{c}{legal (const)} & \multicolumn{2}{c}{near-dom.} \\
 & pt$\to$yrl & yrl$\to$pt & pt$\to$yrl & yrl$\to$pt \\
\midrule
600M (3 seeds) & 31.25 & 35.45 & 64.12 & 56.12 \\
 & \tiny$\pm$0.44 & \tiny$\pm$0.31 & \tiny$\pm$0.38 & \tiny$\pm$0.41 \\
1.3B (1 seed) & 32.03 & \textbf{37.76} & \textbf{66.37} & \textbf{58.22} \\
\bottomrule
\end{tabular}
\caption{Phase-1 evaluation (v1 series): chrF++ for the 600M (mean$\pm$sd over three seeds) and the 1.3B (one run). The 1.3B differences are descriptive: a single run has no seed spread of its own, so the 600M's $\pm$sd does not standardize them.}
\label{tab:phase1}
\end{table}

Phase~1 settles two configuration choices by ablation: related-language mixing and orthography.
Table~\ref{tab:phase1} reports the resulting Phase-1 evaluation on the legal and near-domain holdouts. Mixing
Guarani--Spanish data, the obvious transfer candidate given that our new
token is initialized from Guarani, costs $0.26$ chrF++ and is
discarded: the embedding is a starting point, not a related-language
signal the model can exploit. 

We expected orthographic unification to help, and it did not. Only 29\% of the Constitution's \emph{tokens} (5\% of its types) appear, by exact surface match, in the vocabulary of the largest modern dictionary \cite{avila2021dicionario}; ignoring accents raises these to 46\% and 8\%, which is itself a measure of how much of the divergence is diacritic. Unification nonetheless moved every evaluation set by less than seed noise, with the legal set landing 0.3 chrF++ lower, and it did so with the evaluation references themselves favouring it. We retain a light normalization (punctuation, casing, whitespace) and preserve the original spelling of every pair alongside it. The result is specific to the recipes and data tested here.

Across 8k and 20k training steps and across both model sizes,
legal pt$\to$yrl sits at $\approx$29 chrF++ on dev. Essentially any
configuration lands in the same place, while near-domain translation and
both yrl$\to$pt directions respond normally to training length and model size.
The plateau suggests a data bottleneck specific to article-disjoint legal pt$\to$yrl. \S\ref{sec:analysis} identifies unstable conventional terminology as one important contributor, but the present experiments do not establish that it is the sole cause.

\subsection{Phase 2: terminology-control ablations}
\label{sec:terminology}
\begin{table}[t]\centering\small
\begin{tabular}{lcc}
\toprule
training mechanism & unhinted & hinted \\
\midrule
none (baseline) & 51.5 & not run \\
token pairs only & 55.1 & not run \\
annotation only & 51.0 & 63.6 \\
annotation + token pairs & 54.5 & \textbf{66.2}$\pm$2.7 \\
\bottomrule
\end{tabular}
\caption{TSR on legal development data (198 term occurrences): training mechanism crossed with decode-time hints (v2, 600M). We did not evaluate hints on models trained without annotation; their response to unseen markup would also test robustness to the input format. Single-seed values, except the hinted combination (three-seed mean).}
\label{tab:tsr2x2}
\end{table}
The initial mechanism ablations in Table~\ref{tab:tsr2x2} use 600M models on legal development data for pt$\to$yrl with corpus v2. \emph{Hinted} means glossary terms are supplied inline at decoding; \emph{unhinted} means they are not. Table~\ref{tab:tsr2x2} distinguishes the measured training-only comparison from the within-model effect of hints. Relative to the no-mechanism baseline (28.91 chrF++, 51.5 TSR), the combined mechanism alone, decoded without hints, adds 3.0 pp TSR (54.5, single seed) and 0.99$\pm$0.17 chrF++ over three seeds. On the one seed decoded in both conditions, supplying hints adds a further 8.6 pp (54.5 to 63.1); across three seeds the hinted system reaches 66.2$\pm$2.7 (range 63.1--68.2), 14.7 pp above the unhinted baseline, and brings chrF++ to 1.24$\pm$0.32 above it. The no-annotation hinted baseline was not evaluated, so the interaction between annotation training and hints is not estimated. Annotation alone changes chrF++ by 0.29, within seed variation, and token pairs alone add 3.6 pp unhinted TSR. Hints apply to 110 of 233 segments, so the aggregate effect is bounded by glossary coverage.

\textbf{Negative results:} the LLM-translated Portuguese view of the 925 literary pairs reduces chrF++ by 0.66 in the unhinted condition and by 1.5 in the hinted condition. Because the Portuguese view is added rather than substituted, the experiment jointly changes duplication, source-language weighting, register, and orthographic composition.

In an exploratory failure-targeted distillation experiment on corpus v3,
relative to the annotation-plus-token-pair configuration, we identify
43 terms that fail on legal development data, write 10 legal-register
Portuguese sentences for each (leak-guarded against every evaluation
set) and translate them with the term forced by constrained decoding,
keeping the 233 that survive a length filter. These yield at 600M
(2 seeds) \textbf{+0.93 chrF++ on legal dev} (30.91$\pm$0.25) and
\textbf{+4.2pp unhinted / +7.2pp hinted TSR} (60.4/72.2), with hint chrF++
turning neutral: the glossary migrates into the parameters. A single 1.3B run with the same distilled pairs shows a similar unhinted TSR gain (+4.6pp,
54.5$\to$59.1; single seed) while \emph{halving} the model's dependence
on the hint (hint effect +8.1pp$\to$+4.5pp), though the 600M's chrF++
gain does not transfer (+0.03): at 1.3B the distilled glossary shows up
purely as terminology, invisible to chrF++. Term selection is dev-guided
(declared).

The mechanisms address different parts of the problem: token pairs
expose the model to a term and its spelling, while annotation trains it
to condition on a supplied target form and place it in context,
which is what makes hinting possible at all (51.0 unhinted to 63.6
hinted). Constrained decoding raises the realization rate at inference, but it is lexical enforcement rather than a guarantee of correct terminology.

\subsection{Phase 2: the speech-register gap}
\label{sec:register}
\begin{table}[t]\centering\small
\begin{tabular}{lccc}
\toprule
eval & v2 model & v3 model & $\Delta$ \\
\midrule
speech pt$\to$yrl & 23.06$^{\dagger}$ & \textbf{46.77} & +23.7 \\
speech yrl$\to$pt & 22.82$^{\dagger}$ & \textbf{49.81} & +27.0 \\
legal pt$\to$yrl & 29.90$\pm$0.17 & 29.98$\pm$0.13 & +0.1 \\
near-dom.\ pt$\to$yrl & 60.70$\pm$1.01 & 59.99$\pm$0.64 & $-0.7$ \\
\bottomrule
\end{tabular}
\caption{Effect of the v2$\to$v3 expansion (3{,}408 pairs, 1{,}156 of them speech) on development chrF++. $^{\dagger}$The v2 model saw no speech in training, so the speech rows measure the whole expansion, not the speech subset alone; those two rows are single runs without terminology mechanisms, while the written rows are three-seed means of the shared configuration. The expansion also raises unhinted TSR by 3.0 pp, an effect not attributed to any single source.}
\label{tab:register}
\end{table}
The size of the gap in Table~\ref{tab:register} is worth dwelling on,
because it is easy to misread as a data-volume result. A model trained on
written Nheengatu (legal text, dictionary sentences, school materials) is
not merely worse at spontaneous conversation; at 23 chrF++ its similarity to the
references falls far below the 30 it reaches on the hardest written set
and the 60 on the easiest. Moving from v2 to v3 (an expansion of 3{,}408 pairs, 1{,}156 of them speech, under 9\% of the resulting 13k-pair training set) improves
speech chrF++ by 23.7 to 27.0 points while changing written-domain scores by less than one. The evaluation is speaker-disjoint and session-disjoint, which reduces the risk of speaker memorization. The concentration of the gain on the speech evaluation is consistent with a matched-register effect, but the comparison does not
isolate the speech subset from the other v3 additions. We report this because projects on languages that live primarily in speech are routinely evaluated only on written benchmarks, where a 23-point blind spot is invisible.

\paragraph{What the speech errors look like.} The v2 model's speech outputs were not retained, so we cannot place them beside v3; we characterize the v3 model of Table~\ref{tab:register} instead, on its 116 development turns. The lowest-scoring outputs are one- and two-word turns whose reference depends on discourse context the turn does not carry: \emph{Asu} (`I went') rendered as \emph{vou} (`I go'), \emph{amuyatimana} (`I turned') as \emph{ajunto}. Mid-range outputs are adequate paraphrases that chrF++ scores in the forties: reference \emph{mas da nossa região ou de outra}, system \emph{mas da nossa terra ou outro}. Into Nheengatu the dominant residual is orthographic: \emph{siía pixana ta uiku tatá rendá ruaki} against the reference \emph{siyawa pixana uiku tata rẽda ruaki} differs mainly in accent and nasal marking and scores 42. Two adequacy errors recur: a vocative name rendered as a kinship noun (\emph{Dadá} $\to$ \emph{Papai}, `Dad'), and, in the larger and terminology-augmented models we inspected on the same turns, the reportative \emph{disque} inserted into Portuguese where the source carries no evidential marker.

Register-control codes were validated and then rejected for the final system. A
\texttt{[fala]} tag prepended to conversational sources buys +2.34
chrF++ on speech, a real gain above the noise ruler, but costs a real
$-0.74$ on legal text, and the tagged model collapses to 26 on
speech when the tag is absent, meaning it has learned to depend on a
signal that a deployed system would have to infer. For a single general-purpose model serving mixed traffic, we do not adopt this trade-off. For a dedicated conversational checkpoint, the configuration is promising but requires speaker and user evaluation. Terminology behaviour is unaffected either way,
which is itself informative: register control and terminology control behave as largely separate axes in this experiment.

\subsection{Phase 2: scale at fixed configuration}
\label{sec:scale}
Phase~1 established that model size does not move legal pt$\to$yrl. With
the optimized configuration and the v3 data in place, we can ask a
sharper question: is that a property of the legal domain, or of
generating Nheengatu at all? Holding the configuration fixed and moving
only from 600M to 1.3B, legal pt$\to$yrl lands at 29.67, just below the
600M three-seed spread (29.89--30.12), and spontaneous speech at 46.17,
inside the speech spread (45.79--46.70), while translation \emph{into Portuguese} gains 1.3, 3.0 and 2.5 chrF++ on legal, near-domain and spontaneous speech respectively, and near-domain into-yrl gains 3.1.

Within the tested NLLB family, scaling helps all yrl$\to$pt settings and the easier near-domain pt$\to$yrl setting, but does not resolve the harder article-disjoint legal or spontaneous-speech pt$\to$yrl cases in our runs. Because the 1.3B condition uses a single seed, these differences are descriptive.

The direction gap is consistent with NLLB's own ablations, in which translation into English exceeds translation out of English by a larger margin for very-low-resource directions \cite{nllb2022}. Nheengatu is not explicitly supported by NLLB-200 and has no dedicated pretrained language token.

Hinting continues to help at 1.3B on domain text (+0.55 legal) but costs
$-0.95$ on conversational input, where dictionary-form hints reduce performance. The operational rule we adopt is to hint domain text but not conversation.

\subsection{A same-corpus architecture control}
\label{sec:arch}
We pre-specified a same-corpus control before measuring: full
fine-tuning of Qwen3.5-2B \cite{qwen3.5}, a 2B instruction-tuned
decoder-only model with a hybrid-attention stack, against our 600M
seq2seq, three seeds each, on the development sets. The two sides were not selected alike: the seq2seq kept the checkpoint with the best development chrF++, whereas the decoder-only kept the one with the lowest development loss, which is step 1{,}000 in all three seeds. Under loss-based selection the seq2seq leads all six cells (Table~\ref{tab:arch6}). Because the criteria differ, after submission we also scored the decoder-only's late checkpoints (steps 11--12k, chosen with knowledge of the development scores as a proxy for chrF++-based selection) with the same decoder. The picture changes: the seq2seq still wins conclusively on article-disjoint legal pt$\to$yrl ($+1.90$ chrF++ against a threshold of 1.06 under the pre-specified rule), the decoder-only wins both near-domain cells ($-1.50$ and $-3.23$), and the other three cells are inconclusive. Our declared hypothesis, that the decoder-only would compete into Portuguese, is therefore supported on near-domain yrl$\to$pt once selection is equalized, and not on legal text. Two cautions apply: the late checkpoints are a post-hoc choice, and the near-domain set is where content overlap with training is highest (Limitations), which favours whichever model reproduces training material more readily. This compares \emph{systems}, not architectures (pretraining, size, instruction tuning and checkpoint selection differ together), with one model per side and on development data only, so it does not generalize to decoder-only models as a family. Appendix~\ref{sec:repro} gives both selections, the six-cell table and the hyperparameters.

\subsection{Phase 2: final evaluation}
\label{sec:finaltest}
\begin{table}[t]\centering\small
\setlength{\tabcolsep}{3pt}
\resizebox{\columnwidth}{!}{%
\begin{tabular}{lcccccc}
\toprule
 & \multicolumn{2}{c}{legal} & \multicolumn{2}{c}{near-domain} & \multicolumn{2}{c}{spontaneous speech} \\
system & pt$\to$yrl & yrl$\to$pt & pt$\to$yrl & yrl$\to$pt & pt$\to$yrl & yrl$\to$pt \\
\midrule
flagship & 31.70 & 37.39 & 64.11 & 57.10 & 45.24 & \textbf{49.67} \\
~~+ hints & 31.13 & --- & 64.42 & --- & --- & --- \\
avg & \textbf{32.78} & \textbf{38.14} & \textbf{64.54}$^{\dagger}$ & \textbf{58.76} & \textbf{47.45} & 49.37 \\
\midrule
avg, v1 refs & 32.74 & 38.15 & 62.55 & 56.92 & --- & --- \\
\bottomrule
\end{tabular}}
\caption{Phase-2 final test, decoded once. For pt$\to$yrl, paired bootstrap
intervals do not resolve the difference between our two systems on legal
text (95\% CI $[-0.06,+2.26]$) or near-domain text ($[-0.48,+1.44]$);
the speech interval excludes zero ($[+0.46,+4.06]$). The same holds against Phase 1 on identical
v1 references ($32.74$ vs.\ $32.03$, $[-0.37,+1.79]$, $p=0.20$). The 0.71 gap against Phase 1 on v1 references is not resolved by the paired-bootstrap interval on this 218-segment set. $^{\dagger}$Removing the identified target-side content-overlap items lowers this to 62.09; we report both, and the Limitations give the overlap measurement. }
\label{tab:finaltest}
\end{table}

Table~\ref{tab:finaltest} reports it, for two systems: the
\textbf{flagship} (1.3B, light punctuation/casing/whitespace normalization + annotation + token-pair augmentation, 20k steps) and
\textbf{avg}, the weight average of steps 29k and 30k of a 30k-step
extension of that same run. Note also that the two held-out article sets
differ in difficulty by $\approx$2--3 chrF++ (legal test is the easier
draw), so dev and test numbers are reported separately throughout and
never pooled; the Phase-1$\to$Phase-2 comparisons below use the same evaluation split and reference series. Trivial baselines fix the scale: copying the
Portuguese source scores 13.8 / 12.2 / 9.3 chrF++ on the legal,
near-domain and speech test sets, and shuffled references 13.1 / 9.7 /
11.7. Table~\ref{tab:context} fixes the other end: the spread between our own
three evaluation sets is larger than the spread between published
systems for comparable languages, which is itself the point.

\begin{table}[t]\centering\small
\setlength{\tabcolsep}{3.5pt}
\resizebox{\columnwidth}{!}{%
\begin{tabular}{llcrr}
\toprule
system & language & test set & $\to$Indig. & Indig.$\to$ \\
\midrule
ANLP-25 base & Guarani & XNLI & 35.7 & 35.9 \\
ANLP-25 base & Quechua & XNLI & 31.9 & 37.2 \\
ANLP-25 base & Nahuatl & XNLI & 22.4 & 26.4 \\
ANLP-25 base & Otomí & XNLI & 12.8 & 20.8 \\
\midrule
\textbf{ours 1.3B} & Nheengatu & legal (article-disjoint) & \textbf{32.8} & \textbf{38.1} \\
\textbf{ours 1.3B} & Nheengatu & speech & \textbf{47.5} & \textbf{49.4} \\
\textbf{ours 1.3B} & Nheengatu & near-dom.\ & \textbf{64.5} & \textbf{58.8} \\
\bottomrule
\end{tabular}}
\caption{Context from AmericasNLP, not a ranking: no cell is a like-for-like comparison. All values are chrF++ from fine-tuned NLLB-200-distilled-1.3B models with an extended embedding matrix; our rows are the \emph{avg} system of Table~\ref{tab:finaltest}. The AmericasNLP pairs use Spanish and translated XNLI test sets; ours use Portuguese and single-step translations (field translation on the speech side, see Limitations).}
\label{tab:context}
\end{table}

The legal and near-domain sets were first evaluated in Phase 1 and reused
in this evaluation, so we treat them as reused holdouts rather than
pristine single-use test sets. Each Phase-2 model--test combination was
decoded once, with no test-based checkpoint selection. The speech test
set remained untouched until this pass. Before decoding we fixed, in writing, which checkpoint would be evaluated, which metric would decide, and what each possible outcome would mean.

First, the ranking flipped between dev and test. On dev the flagship
checkpoint beat the checkpoint-averaged system; on test the average won
the headline metric by more than a point on the legal set. Had we
selected on dev we would have promoted the system test ranked second;
had we selected on test we would have been fitting it. We report both,
which quantifies the selection risk a 218-segment set carries.

Second, cross-series scoring punishes cleaner models: scored against the
\emph{original} v1 references our v3 systems reach 62.6 on near-domain
against 66.4 for the v1-trained system, because a model trained not to
emit the artefacts those references still contain is penalised for their
absence. We therefore never compare across reference series.

Third, the closed baseline \cite{goncalves-etal-2025-exploring} cannot be
compared to ours directly. Rescoring our outputs in \emph{their} protocol gives 64.48 pt$\rightarrow$yrl and 60.2
yrl$\rightarrow$pt on near-domain against their 69.1 and 66.9. What matters more is that their test set is a random split of the
corpus the model was trained on (6{,}699 training pairs, 233 test pairs,
randomly divided); the same paper splits its other language by source
instead, and reports 40.2 there. A random split of the training corpus and a held-out test of unseen legal prose measure different things, and only the second probes unseen legal content.

The test also reproduces the terminology dissociation. Without
any inference-time intervention, the Phase-2 system scores 47.8 TSR
against the Phase-1 system's 44.0 on the same 232 term occurrences
(\textbf{+3.8pp}): data and training alone internalised part of the
glossary. Adding hints then moves TSR from 47.8 to
57.3 (+9.5pp) while costing 0.57 chrF++, the same dissociation observed on development data, reproduced under the pre-specified Phase-2 evaluation. The averaged system, which obtains the highest chrF++ of our two final systems, simultaneously obtains the lower terminology score, at 44.4 TSR.

\subsection{Decoding without a validated neural metric}
\label{sec:decoding}
\begin{figure}[t]\centering
\includegraphics[width=\columnwidth]{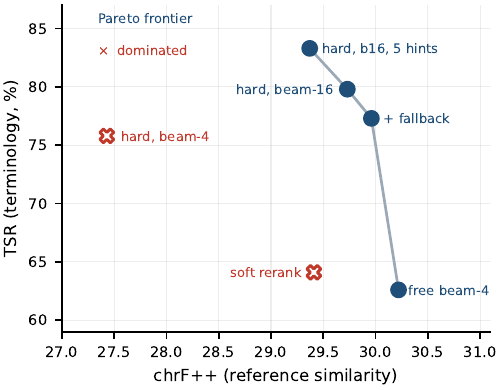}
\caption{Terminology realization versus chrF++ on hinted legal development data.
Free beam-4 gives 30.22 chrF++ at 62.6 TSR; hard constraints with a wide
beam and five hints reach \textbf{83.3} TSR at 29.37. Two settings are
\emph{dominated} in both dimensions at once: soft reranking (29.41/64.1) and hard
constraints at beam-4 (27.43/75.8), where narrow constrained search substantially reduces reference similarity and a wider beam recovers part of the loss, as the
dynamic-beam-allocation literature predicts
\cite{post-vilar-2018-fast}. The decoder realizes 164 of 165 requested surface-form insertions, or 99.4\%. This measures surface inclusion rather than semantic, morphological, or syntactic correctness. The wide-beam gain is hint-conditional:
the argmax drops only when hints are present. }
\label{fig:frontier}
\end{figure}

Because no learned quality metric is validated for yrl, we do not use neural-metric utility as primary evidence for reranking. We instead evaluate search variants, string-based MBR, terminology reranking, and lexical constraints. Five are null. Three
are measured on the near-domain dev set against a beam-4 reference of
63.1 chrF++: sampled chrF-MBR (61.9--62.6, even with the beam candidate
in the pool); MBR reranking of the beam's own n-best (damages the
argmax); and wider beams, which saturate unhinted. The other two are
conservative LLM post-editing of the Portuguese side ($-0.2$ chrF++; no improvement in reference similarity) and soft terminology reranking, dominated on the
legal-dev frontier below. One frontier does validate
(Figure~\ref{fig:frontier}).

\paragraph{A cost chrF++ almost hides: constraints multiply
degeneration.} Repetition detection over the same 233 outputs
(punctuation-normalized $n$-grams; \textbf{zero false positives on the
233 human references}, and no source sentence contains list markers)
flags 1.7\% of free beam-4 outputs and 1.3\% of hinted ones, against
6.0\%, 7.7\% and 8.2\% for the three constrained settings (paired
uncorrected exact two-sided McNemar $p=.031/.003/.0015$; the ratio holds at 3.2--4.8$\times$ across
four detector thresholds). The failure localizes to the constraints
themselves: constrained runs consume the \emph{plain} source (terms enter
only through \texttt{force\_words\_ids}), and among the 110
sentences that actually receive constraints degeneration rises from
3.6\% to 16.4\%, while constraint-free sentences under the same decoder
stay at 0.8\%. Every flagged output is a genuine loop on inspection
(median length ratio 3.2, up to 17.8$\times$ the reference, sometimes
fabricating enumerations absent from the source).

chrF++ does penalize these sentences ($-7.2$ each against the same sentences decoded freely), but its recall-weighted $F_{\beta=2}$ makes appended repetition cheap: a hypothesis consisting of the reference \emph{tripled} still scores 71.0. Corpus-level the whole pathology is therefore worth only $-0.59$.

A repetition-triggered fallback (detect the loop, re-decode that sentence without constraints) buys it back. It fires on 19 of 233 sentences and gives the fallback point of Figure~\ref{fig:frontier}: 29.96 chrF++ (within 0.26 of unconstrained
decoding), degeneration down from 8.2\% to 0.9\%, and TSR at 77.3. That
last figure is the honest cost: the fallback trades $-6.0$pp of
terminology for $+0.59$ chrF++, because the rescued sentences lose their
forced terms by construction. Whether that trade-off suits a community
writing assistant must be settled by speaker and user evaluation.

\section{Analysis}
\label{sec:analysis}
\textbf{chrF++$\times$TSR dissociation.} Scale buys chrF++ without
terminology (test TSR flat at 600M$\to$1.3B in Phase 1); annotation + token-pair augmentation buys both; extended training buys chrF++ \emph{and loses}
terminology (the avg system). Per-case verification of all 34
hint-induced term flips shows 24/26 gains syntactically integrated and
only 1 real regression: the matcher-based TSR \emph{understates} the
mechanism.

\textbf{Register-conditional orthography switching:} trained on multiple
traditions, the model in qualitative inspection sometimes follows the
orthographic tradition associated with the input register; we do not
quantify this effect, and we propose per-norm control codes as the
community-governed mechanism.

\paragraph{Terminology errors on article-disjoint legal text.} We inspect both model sizes on the same 20 legal-development segments, 40 outputs in all. Outputs generally preserve grammatical frames, legal citations such as \emph{art.\ 89, VII}, and many numerical expressions, while much of the residual error involves conventional legal terms such as \emph{civil registry}, \emph{death certificate}, and \emph{cruel} (from \emph{penas cruéis}, `cruel punishments', art.\ 5, XLVII). The 90 missed term occurrences correspond to 60 distinct glossary forms; only eight occur at most twice in training, and the median training count is 10. Simple token rarity is therefore not the only explanation. Several terms have competing or inconsistent realizations across the written sources we examined. We treat this instability as an important bottleneck, but the present sample does not establish that terminology accounts for the entire legal-performance plateau.

Two further observations from the same samples. First, in $\ge$3/20 legal samples the system is \emph{more faithful to the source than the human reference}: it renders ``sixty'' where the reference prints ``65'', and ``two rounds'' where the
reference has ``morning or afternoon''. chrF++ against such references
has a ceiling well below 100, which is why our speaker protocol asks evaluators to judge against the \emph{source}. Second, the hint
mechanism's own failure modes are narrow but real: noun hints pasted
into verbal slots, and a digit juxtaposed with the hinted native numeral (\emph{``10
ára-ita mukũi pu aára''}, literally ``10 days two hands''); conversely,
in two cases a hint \emph{cured} a degeneration loop.

\section{Discussion}
Article-disjoint legal pt$\to$yrl plateaus across the tested training lengths and model sizes. Our error analysis points to unstable conventional terminology as an important contributor: the largest modern yrl dictionary \cite{avila2021dicionario} provides a usable match for only one of 54 outstanding legal terms, and the tested
increase from 600M to 1.3B did not resolve terms that are inconsistently attested in the written resources we examined.

The analysis has a practical implication. For languages whose written resources do not consistently attest specialized terminology, a community-governed control interface can complement the model: a revisable glossary, mechanisms that expose the model to it during training, and an optional lexical-enforcement mode at inference time. The hint budget matters as well: allowing four inline hints per sentence at inference, instead of the two used in training, raises TSR by 1.9$\pm$0.3 pp on the v3 600M models (three seeds) at no chrF++ cost, because sentences that match more than two glossary terms otherwise lose the rest at the cap. The decoding mode increases specified surface-form realization but does not guarantee semantically, morphologically, or syntactically correct use. The components are independently replaceable, allowing the pipeline to incorporate an authoritative terminology list if and when one is established.

Within the tested NLLB family, increasing model size is consistently helpful for yrl$\to$pt and also improves near-domain pt$\to$yrl, but it does not improve article-disjoint legal or speech pt$\to$yrl in our runs. A larger checkpoint is therefore promising for reading support, whereas writing support in the harder registers is likely to require better matched data, terminology governance, and user evaluation in addition to model size.

Finally, the three dissociations we report (chrF++ against terminology,
chrF++ against degeneration, dev ranking against test ranking) all point
the same way: for languages without a validated learned metric, a single automatic
number gives a projection of quality along one axis, unstable at this
sample size, and never a summary of it.
The next required instrument is speaker-grounded evaluation of adequacy, naturalness, terminology, and usefulness. The planned protocol is described in the Ethics Statement, but no speaker-evaluation result is included in the present version.

\section*{Limitations}
\textbf{Limited architectural coverage.} Our principal systems are
NLLB-200 fine-tunes with an added language token. \S\ref{sec:arch} reports a
same-corpus control against a 2B decoder-only system, but we did not
train an alternative encoder-decoder base such as mBART or M2M-100, and
the control itself covers one model and one scale per side, on
development data only.

\begin{itemize}\itemsep2pt \parskip0pt
\item Small evaluation sets (218--372 per set; speech 115--116). At this
size differences of about a point were not resolved by our bootstrap intervals
(Table~\ref{tab:finaltest}).
\item No speaker evaluation is included in this version.
\item chrF++ penalizes valid cross-orthography outputs.
\item TSR synonym coverage is incomplete, and shown to understate.
\item The 1.3B models are single-seed.
\item Failure-targeted synthetic term selection is development-guided, and no independent test result is included. We therefore treat the distillation result as exploratory.
\item Brazilian text-and-data-mining (TDM) legality for third-party research use lacks an explicit exception. Our posture is non-commercial research; one third-party layer is redistributed with its author's written permission and the others remain pending.
\item Domain status of the legal evaluation. The legal development and test sets hold out complete constitutional articles, but they come from the same translated Constitution as 2,031 training pairs. They are article-disjoint and segment-disjoint, not domain-disjoint or document-disjoint.
\item Terminology governance. The glossary is source-derived and researcher-adjudicated rather than community-authorized. Its forms should be treated as provisional alternatives for experimentation, not as an official Nheengatu terminology standard. Its reach is also bounded: in the legal test set, 8.7\% of content tokens (176 word types) never occur in training (by surface form) and none has a glossary entry, so no glossary-based mechanism can act on them.
\end{itemize}

\paragraph{Content overlap in the near-domain test set.} Exact-pair
deduplication is not sufficient when a language has few sources. Our
near-domain test contains no leaked pair (zero identical source sentences,
zero identical targets, zero identical pairs against training), but 23 of
its 371 targets share an 8-gram with training data,
because the same traditional narratives appear in both the dictionary
and the treebank, in different renderings. Those 23 items score 80.5
against the set's 64.5, and removing them lowers the set to
\textbf{62.1}. We report both: 64.5 for the declared set and 62.1 as the
conservative reading. The legal and speech sets are practically unaffected (1/218
and 0/115 targets, respectively). The development sets show the same
pattern (3/233 legal, 25/372 near-domain, 0/116 speech), which matters
because the mechanism and decoding results are measured there. This
overlap reflects a structural challenge in low-resource corpus
construction, and content-level $n$-gram overlap should be reported
alongside exact pair-level deduplication. On the Portuguese source side the same measurement returns 21 of 371
near-domain and 27 of 218 legal segments; the legal figure reflects
formulaic legal phrasing recurring across disjoint articles rather than
content leakage, and it is the mirror image of the yrl side, where
orthographic variation suppresses exact recurrence.

\paragraph{What our speech numbers are computed over.} Two properties of
the time-aligned speech source qualify the speech scores we report.
First, its alignment unit is the pause-delimited \emph{turn}, not the
sentence (median 1.8\,s and 4 words, a third of them two words or fewer;
only 10\% end in sentence-final punctuation), so most segments are mid-utterance fragments, a unit at once
easier than a sentence (short) and harder (fragmentary,
context-dependent). Second, its Portuguese side is a field
free-translation written by the transcribers: regional Upper Rio Negro
and/or L2 Portuguese (sentence-final discourse \emph{nem} in 13.0\% of
the speech test set, non-standard spelling 9.6\%, non-standard agreement
2.6\%; the constitutional set scores 0\% on all three). Into Portuguese,
some model outputs use standard Portuguese forms where the
references use regional or non-standard forms, which chrF++ can
penalize sentence by sentence: \emph{``Lili,
aonde está''} (reference) vs.\ \emph{``Lili, onde está''} (system) scores
64.3, and 100.0 against a standardized reference. Aggregated the effect
vanishes (standardizing unambiguous spellings moves 12 runs by $-0.61$ to
$+0.62$, mean $\approx-0.07$), so no headline number needs correcting; we
leave the references untouched, which is also the CARE-consistent choice:
``correcting'' Indigenous speakers' Portuguese is not ours to do. In the pt$\rightarrow$yrl direction the same text is \emph{input};
robustness to standardized Portuguese input remains untested.

\section*{Ethics Statement}
Our process is informed by the ethical framework of \citet{mager-etal-2023-ethical} and by CARE principles. We contacted the Academia da Língua Nheengatu but have not obtained a response, endorsement, or institutional approval; silence is not treated as consent. The glossary and orthographic choices in this work are therefore provisional and source-derived rather than authoritative. Future governance should involve speakers from multiple regions and orthographic traditions. Licensing is layered: official-text layer outside copyright protection under Article 8 of Brazilian Law 9,610/1998; the treebank and speech layers inherit CC BY-NC-SA; the dictionary-example layer \cite{avila2021dicionario} is redistributed with the author's written permission; the remaining third-party layers ship as reproducible extraction pipelines pending permission; the three NLLB-derived checkpoints are publicly available on Hugging Face under CC-BY-NC-4.0; the Qwen-derived checkpoint is released under Apache-2.0. We position the system as writing/reading
support, never official translation. 

\paragraph{Speaker evaluation protocol.} No speaker evaluation has been run for this version; the instrument below is built and being fielded, and its results will be reported separately. It presents 75 items in a self-contained offline form: absolute adequacy/naturalness on 30 system outputs balanced across legal, near-domain and speech registers, interleaved with \textbf{9 hidden human-reference anchors} (23\% of the block, WMT-style) that give a calibration ceiling and an
inattentive-rater check; 20 blind pairwise items (free vs.\ constrained
decoding, with preference asked \emph{before} the target term is
revealed); 12 blind system-vs-official-translation items; and 4 open
language questions. Item counts follow a power analysis. Speech items carry
their neighbouring conversational turns as explicitly non-evaluated
context, and evaluators are told the Portuguese of those turns is the
transcribers' own regional writing, not a translation error. Evaluators
are told translations come from undisclosed mixed sources, never that all
are machine-made, and are \textbf{debriefed within the form}
about the hidden human references and why they were concealed; blinding
keys never ship with the questionnaire. Evaluation data are anonymous by default. Any option to be acknowledged by name is collected through a separate, explicit opt-in that is not linked to the participant's individual ratings. Participation is voluntary and compensated, with informed consent obtained before rating and the right to withdraw data until analysis. Ratings are stored under pseudonymous identifiers with restricted access. No speaker data have been analysed for this submission.

\section*{Acknowledgments}
We thank Marcel Twardowsky Avila for permission to redistribute the dictionary example pairs derived from his thesis; the Revista Leetra Indígena at UFSCar and the authors of its bilingual school books; and the fifteen Indigenous translators of the Upper Rio Negro and Middle Tapajós whose translation of the Constitution, published by the STF and the CNJ, anchors this corpus.

\bibliography{refs}

\appendix
\section{Reproducibility}
\label{sec:repro}

\paragraph{Hardware and budget.} A 20k-step 600M run takes 72--77 minutes
on a single NVIDIA H100 ($\approx$1.3 GPU-h); the 1.3B flagship needs
about four hours of wall clock. Jobs killed by a wall-clock limit resume from the last checkpoint, and we note one pitfall we hit: early stopping
restarts its patience counter on resume, so patience is not comparable
across interrupted and uninterrupted runs.

\paragraph{Training.} Base checkpoints are NLLB-200-distilled 600M and
1.3B. The new language token \texttt{yrl\_Latn} is initialized from the
\texttt{grn\_Latn} (Guarani) embedding, a starting point only; mixing
Guarani training data was measured and rejected ($-0.26$ chrF++).
Hyperparameters follow AmericasNLP submissions in the same line
\cite{gowsmith2023sheffield,garcia-gilabert-etal-2024-bsc}: full
fine-tuning (no adapters), lr $1\times10^{-4}$ with inverse-square-root schedule and 400 warmup steps, weight decay 0.01, no label smoothing,
max sequence length 192, effective batch 32 (16$\times$2 accumulation),
bf16, 20k steps, evaluation every 1000 steps, best checkpoint selected by
dev chrF++ with early-stopping patience 8, beam 4 for the in-training
generation.

\paragraph{The architecture control.} The decoder-only side of
\S\ref{sec:arch} is \texttt{Qwen/Qwen3.5-2B} \cite{qwen3.5}, the
instruction-tuned release rather than the pretrained base, trained on
the same v3 data, with the same three seeds, and scored with the same
decoding and metrics as the seq2seq runs. The original Qwen checkpoint is a
vision--language foundation model whose card counts 2B parameters in the
language model; we load it with the text-only
\texttt{Qwen3\_5ForCausalLM} class, which instantiates the language
backbone alone (1.882B parameters, the vision tower never constructed) and we use its own tokenizer (248{,}320 pieces, tied input and output
embeddings). Adaptation is full fine-tuning with no adapters: lr
$2\times10^{-5}$ with cosine schedule and 400 warmup steps, weight decay
0.01, fused AdamW, fp32 master weights under bf16 autocast, gradient
checkpointing, maximum sequence length 512 for prompt and target
together, effective batch 32 ($8\times4$ accumulation), 20k maximum
steps, evaluation every 1000 steps on a 600-example development subset,
best checkpoint by development loss (the seq2seq side keeps the checkpoint with the best development chrF++; Table~\ref{tab:arch6} scores both selections). Training pairs are rendered with a plain Portuguese instruction template rather than the chat template (\emph{Traduza do <source> para o <target>.}, then the source line and an empty target line), and the reasoning-delimiter tokens are suppressed at
decoding so that generation budget cannot be spent on them. In all three
seeds the best checkpoint is step 1{,}000 (1.15 epochs over the
27{,}900 bidirectional training examples) and the remaining 19k steps never improve development loss. Development chrF++ nonetheless keeps improving well past that point (Table~\ref{tab:arch6}): at steps 11--12k the three seeds reach 27.67$\pm$0.57 on legal pt$\to$yrl against 25.89$\pm$0.67 at step 1{,}000, and 62.37$\pm$0.54 against 52.11$\pm$0.82 on near-domain pt$\to$yrl. The conclusiveness threshold is three standard errors of the difference between the two three-seed means, computed separately for each set,
direction and metric; across the six cells of the loss-selected comparison it ranges from 0.98 to 3.73 chrF++. Speech yrl$\to$pt clears it by 0.06 chrF++ but falls 0.19 short under spBLEU, which is why the cell is reported and not claimed.

\begin{table}[t]\centering\small
\setlength{\tabcolsep}{3.5pt}
\resizebox{\columnwidth}{!}{%
\begin{tabular}{lcccrr}
\toprule
cell & seq2seq 600M & dec.-only @1k & dec.-only late & $\Delta$ late & $3\cdot$SE \\
\midrule
legal pt$\to$yrl & 29.57$\pm$0.22 & 25.89$\pm$0.67 & 27.67$\pm$0.57 & $+1.90$ & 1.06 \\
legal yrl$\to$pt & 34.83$\pm$0.29 & 28.14$\pm$0.68 & 33.11$\pm$1.51 & $+1.72$ & 2.67 \\
near-dom.\ pt$\to$yrl & 60.87$\pm$0.49 & 52.11$\pm$0.82 & 62.37$\pm$0.54 & $-1.50$ & 1.26 \\
near-dom.\ yrl$\to$pt & 56.68$\pm$0.23 & 48.94$\pm$0.51 & 59.91$\pm$0.05 & $-3.23$ & 0.41 \\
speech pt$\to$yrl & 46.23$\pm$0.48 & 39.96$\pm$1.11 & 44.74$\pm$0.90 & $+1.49$ & 1.77 \\
speech yrl$\to$pt & 50.95$\pm$1.05 & 47.16$\pm$1.88 & 47.88$\pm$1.60 & $+3.08$ & 3.31 \\
\bottomrule
\end{tabular}}
\caption{The architecture control under both checkpoint selections (development sets, chrF++, mean$\pm$sd over three seeds). ``@1k'' is the loss-selected checkpoint; ``late'' is steps 11--12k, scored after submission as a proxy for chrF++ selection. $\Delta$ is seq2seq minus late decoder-only (positive favours the seq2seq); $3\cdot$SE is three standard errors of the difference, the pre-specified threshold for that comparison, and a cell is conclusive when $|\Delta|\ge 3\cdot$SE and $|\Delta|\ge 1.0$: legal pt$\to$yrl (seq2seq) and both near-domain cells (decoder-only).}
\label{tab:arch6}
\end{table}
\paragraph{Software.} Metrics are computed with sacreBLEU 2.6.0
(\texttt{nrefs:1}, \texttt{case:mixed}; chrF++ with \texttt{nc:6},
\texttt{nw:2}, \texttt{space:no}; spBLEU with \texttt{tok:flores200});
NLLB models are trained with Transformers 4.55.4 and PyTorch 2.6.0,
seeds 1--3. The Qwen control uses a separate environment with
Transformers 5.14.1. Confidence intervals use paired bootstrap resampling with 1{,}000
resamples.

\paragraph{Evaluation.} chrF++ \cite{popovic-2017-chrf} is sacreBLEU's \cite{post-2018-call} \texttt{CHRF} with \texttt{word\_order=2} and the default $\beta=2$. Table~\ref{tab:metrics} reports the full test results additionally in plain chrF
(\texttt{word\_order=0}) and spBLEU (\texttt{tokenize=flores200}), so
that they can be read against work using either convention: much of the
Indigenous-language literature reports plain chrF, and spBLEU is the
severest of the three on a morphologically rich target. We deliberately
use \emph{no} neural metric: COMET, CometKiwi
and MetricX have not been validated for yrl and provide no dedicated
language coverage, so their scores here would be uninterpretable. TSR (\S\ref{sec:setup}) is scored over
the union of the audited glossary and the base bilingual lexicon, with a
yrl lemma table, a fuzzy threshold of 0.85 applied only to forms of five
characters or more, and single-word Portuguese terms under four
characters skipped. Each model--test combination was decoded once, under a written protocol fixing the system, the checkpoint and the metric before decoding. The legal and near-domain sets are reused holdouts, first evaluated in Phase 1 and once in Phase 2; \texttt{test\_fala} was spent in the Phase-2 pass.

\begin{table}[t]\centering\small
\begin{tabular}{llccc}
\toprule
set & dir. & chrF++ & chrF & spBLEU \\
\midrule
\multicolumn{5}{l}{\emph{flagship}} \\
legal & pt$\to$yrl & 31.70 & 35.95 & 12.83 \\
legal & yrl$\to$pt & 37.39 & 39.56 & 19.90 \\
near-dom. & pt$\to$yrl & 64.11 & 65.99 & 54.65 \\
near-dom. & yrl$\to$pt & 57.10 & 57.86 & 41.68 \\
speech & pt$\to$yrl & 45.24 & 49.77 & 27.86 \\
speech & yrl$\to$pt & 49.67 & 50.88 & 31.71 \\
\midrule
\multicolumn{5}{l}{\emph{avg}} \\
legal & pt$\to$yrl & 32.78 & 36.93 & 13.43 \\
legal & yrl$\to$pt & 38.14 & 40.29 & 20.73 \\
near-dom. & pt$\to$yrl & 64.54 & 66.44 & 54.85 \\
near-dom. & yrl$\to$pt & 58.76 & 59.57 & 43.49 \\
speech & pt$\to$yrl & 47.45 & 51.89 & 30.56 \\
speech & yrl$\to$pt & 49.37 & 50.66 & 31.57 \\
\bottomrule
\end{tabular}
\caption{Final test under three metrics. Plain chrF is approximately 0.8--4.5 points
above chrF++ in these results; chrF++ additionally incorporates word
$n$-grams. spBLEU scores are lower on these sets. The ordering of the
two systems within each set and direction is the same under all three
metrics; comparisons across directions need not have the same ordering.}
\label{tab:metrics}
\end{table}

\paragraph{Audit categories and residual estimates.} Two v3 sources were
audited on samples of 200 and 240 pairs, with evidence attached to every judgement:
200 speech pairs, scored against the temporal context of neighbouring
turns, and 240 dictionary-example pairs, scored against a window of the
page's source text.

\begin{table}[ht]\centering\small
\setlength{\tabcolsep}{4pt}
\begin{tabular}{lrrrr}
\toprule
 & ok & minor & suspect & defective \\
\midrule
Speech (200) & 183 & 11 & 2 & 4 \\
Dictionary ex.\ (240) & 208 & 16 & --- & 16 \\
\bottomrule
\end{tabular}
\caption{Audit outcomes; the approval rates quoted in
\S\ref{sec:corpus} are the \emph{ok} column.}
\label{tab:audit}
\end{table}

The residual estimates are lower than the non-ok mass in
Table~\ref{tab:audit} because the two tails were treated differently. In
the speech sample all four defective pairs were code-switch identity
pairs, a mechanically detectable class that we then removed
population-wide (42 pairs across the full resource); what survives is the
suspect category, 2/200 = 1.0\% (Wilson 95\% CI $[0.3, 3.6]$). In the
dictionary sample the 31 corrections returned by the audit were verified
against the page sources and applied to the audited pairs only, so the
observed defective fraction is 16/240 = 6.7\% (Wilson interval
$[4.1, 10.6]$). Because the sampling procedure was not preserved,
these sample proportions and nominal intervals do not establish
population error bounds for the unaudited pairs. Both residual classes are distinct
from mispairing: 28 of the 32 dictionary problems are left-edge
truncation or bleed on the yrl side, the Portuguese side is correct in
239/240, and cross-pairings are zero.

\paragraph{The French--yrl pairs.} The 925 FR--yrl pairs are concatenated
with the pt--yrl training file before the bidirectional expansion, so
both directions are trained: 925 fra$\to$yrl and 925 yrl$\to$fra,
1{,}850 of the 27{,}900 training examples, and 12{,}467 yrl running
words, 9.4\% of the yrl side the model is fit to. They carry NLLB's own
\texttt{fra\_Latn} and \texttt{yrl\_Latn} tags and no additional source
or register tag. There is no upsampling, downsampling or per-source
balancing: the concatenated set is shuffled once, so every source
contributes in proportion to its size. Inline terminology annotation is
applied only to \texttt{por\_Latn} sources, so the French pairs are never
annotated. yrl$\to$fra is trained and never evaluated; the mirroring is
applied uniformly to every training file rather than decided per source.

\paragraph{Glossary provenance.} The 904 training-glossary entries were
mined from the training pairs and from external lexical resources.
Evaluation references enter the pipeline only as an exclusion filter,
deleting mined candidates that coincide with a development or test pair;
they are never a source of entries. Checking every entry against the
Portuguese side of both partitions, none has support only in the
evaluation references.

\paragraph{Statistics.} Beyond the $\pm0.35$ chrF++ ruler of
\S\ref{sec:setup}, the TSR rulers on the legal dev set are $\pm1.6$pp
unhinted and $\pm2.7$pp hinted (3 seeds, 600M): a single-run TSR
difference below $\sim$3pp is not interpretable, and we mark such
comparisons as inconclusive instead of reporting them as effects. Of the
two single-seed results that failed to replicate, one was a TSR. Paired
comparisons of degeneration rates use McNemar's exact test. The metric values covered by the recorded audit rounds were recomputed
from stored hypothesis files independently of the training-time logs.
The v2 speech scores are retained evaluation summaries; their hypotheses
were not preserved. The released code includes the terminology matcher
and training glossary. The base bilingual lexicon, the scoring lexicon's
other component, contains sentence-length entries from sources whose
redistribution is still pending.

\paragraph{Code, data and models.} Four checkpoints are released on
Hugging Face: the NLLB 600M annotation-plus-token-pair model (seed 3),
the 1.3B flagship and checkpoint average, and the text-only Qwen model
(seed 2, step 11{,}000). The Qwen release has the highest near-domain
pt$\to$yrl development chrF++ among the three late checkpoints.
Table~\ref{tab:arch6} reports three-seed architecture-control results;
its NLLB 600M comparator uses no terminology mechanisms. The final-test
results in Table~\ref{tab:finaltest} concern the flagship and average.
Model cards and weights are linked at
\url{https://huggingface.co/voz-ancestral}.
The extraction, alignment, annotation,
constrained-decoding and evaluation scripts, and the
corpus layers whose licences permit redistribution are at
\url{https://github.com/fabriciocarraro/NheengatuMT}.
Third-party layers ship as reproducible extraction
pipelines instead of redistributed text where permission is pending; the
dictionary-example layer is included with its author's written permission.

\end{document}